\documentclass[runningheads]{llncs}
\usepackage[T1]{fontenc}
\usepackage{graphicx,verbatim}
\usepackage{booktabs}  
\usepackage{array}
\usepackage{amsmath}
\usepackage{amssymb}
\usepackage[
  separate-uncertainty = true,
  multi-part-units = repeat
]{siunitx}

\DeclareMathOperator*{\argmin}{argmin}
\begin{document}
%
\title{Neural Autoregressive Transformers \\for Modeling Brain Aging}
\title{Neural Autoregressive Modeling of Brain Aging}
%

\author{Ridvan Yesiloglu\inst{1} \and
Wei Peng\inst{2} \and
Md Tauhidul Islam\inst{3} \and
Ehsan Adeli \inst{2,4,5}}
\authorrunning{R. Yesiloglu et al.}
%
\institute{Department of Electrical Engineering, Stanford University, Stanford, CA, USA  \and
Dept. of Psychiatry \& Behavioral Sciences, Stanford University, Stanford, CA, USA \\ 
\and Department of Radiation Oncology, Stanford University, Stanford, CA, USA  
\and Department of Biomedical Data Science, Stanford University, Stanford, CA, USA \\
\and Department of Computer Science, Stanford University, Stanford, CA, USA
\email{ridvan@stanford.edu}}

    
\maketitle              
\begin{abstract}
Brain aging synthesis is a critical task with broad applications in clinical and computational neuroscience. The ability to predict the future structural evolution of a subject’s brain from an earlier MRI scan provides valuable insights into aging trajectories. Yet, the high-dimensionality of data, subtle changes of structure across ages, and subject-specific patterns constitute challenges in the synthesis of the aging brain. To overcome these challenges, we propose NeuroAR, a novel brain aging simulation model based on generative autoregressive transformers. NeuroAR synthesizes the aging brain by autoregressively estimating the discrete token maps of a future scan from a convenient space of concatenated token embeddings of a previous and future scan. To guide the generation, it concatenates into each scale the subject’s previous scan, and uses its acquisition age and the target age at each block via cross-attention. We evaluate our approach on both the elderly population and adolescent subjects, demonstrating superior performance over state-of-the-art generative models, including latent diffusion models (LDM) and generative adversarial networks, in terms of image fidelity. Furthermore, we employ a pre-trained age predictor to further validate the consistency and realism of the synthesized images with respect to expected aging patterns. NeuroAR significantly outperforms key models, including LDM, demonstrating its ability to model subject-specific brain aging trajectories with high fidelity. 


\end{abstract}
\section{Introduction}

Brain aging is a complex and heterogeneous process influenced by genetic, environmental, and pathological factors. Understanding the structural evolution of the brain over time is crucial for characterizing aging trajectories \cite{Ziegler_et_al}. Longitudinal brain imaging studies provide valuable insights into these processes, yet acquiring follow-up MRI scans is expensive, time-consuming, and subject to participant dropout. Consequently, the ability to synthesize realistic aging brain MRI scans has significant implications for modeling, early diagnosis, and data augmentation in deep learning applications.

The synthesis of brain aging involves approaches categorized into population-level and subject-level generation. Traditional methods, such as patch-based dictionary learning \cite{zhang2016patch}, kernel regression \cite{huizinga2018kernel,serag2012kernel}, linear mixed-effect modeling \cite{lorenzi2015linear,sivera2019linear}, and non-rigid registration \cite{sharma2010nonrigid,modat2014nonrigid}, have been used to create spatio-temporal atlases of the brain. However, these population-based approaches often fail to capture subject-specific trajectories due to their reliance on average data.  To address this limitation, deep generative models have been employed to learn subject-specific brain aging trajectories. For instance, CounterSynth \cite{pombo2023equitable} was introduced as a GAN-based counterfactual synthesis method that simulates various conditions in brain MRIs, including aging, by generating diffeomorphic transformations that reflect specified covariates. However, due to the computational burden, most of such methods rely on 2D generation techniques \cite{xia2019brainage,ravi2019daninetmiccai}, and 3D generation that depends on consistency-enforced 2D outputs \cite{ravi2022degenerative}, which often exhibit limited temporal consistency. 

Building on the success of diffusion models, Yoon et al. proposed a diffusion model conditioned on a sequence of scans to synthesize longitudinal images \cite{yoon2023sadm}. Puglisi et al. introduced Brain Latent Progression (BrLP) \cite{puglisi2025brainlatent}, which is based on a Latent Diffusion Model \cite{rombach2022latentdiffusion}. While this method shows impressive accuracy, it is limited by high sampling times due to the iterative denoising steps inherent in diffusion models.


To overcome the limitations of diffusion models and GANs and to transfer the recent promising success of the autoregressive models (AR) in the language domain \cite{gpt_series_3,ar_llms_1,ar_llms_9_team2023gemini} to image generation, recently, Tian et. al. \cite{VAR} introduced visual autoregressive model (VAR), an autoregressive transformer model on the latent domain of a VQVAE \cite{vqvae}. VAR organizes the latent representation of an image into multiple scales and works by autoregressively predicting the next scale of tokens. VAR starts with a special start token, specifically a category embedding for image generation, and produces 256x256 images.

NeuroAR synthesizes the aging brain by autoregressively estimating the discrete token maps of a future scan from a convenient space of concatenated token embeddings of a previous and future scan. To guide the generation, it concatenates into each scale the subject’s previous scan, and uses its acquisition age and the target age at each block via cross-attention.

In this work, we introduce NeuroAR (Neural Autoregressive Model), a novel brain aging simulation model which employs a scale-wise autoregressive generation paradigm. NeuroAR forms a convenient longitudinal space of embeddings by concatenating the token embeddings of a previous and a next scan. From this space, transformer blocks autoregressively generate the discrete token maps of the next scan from coarse to fine scales, which are then fed into a decoder to predict the next scan. To guide the generation, NeuroAR embeds the acquisition age and the target age into each scale via cross-attention and adaptive normalization layers.

We evaluate our approach on the ADNI, PPMI, and ABCD datasets, demonstrating superior performance over state-of-the-art generative models, including the latent diffusion model (LDM) and generative adversarial network (GAN), in terms of image fidelity on both aging adults and children. Quantitative assessments using standard image quality metrics confirm that our method generates significantly more accurate representations of brain aging compared to ground-truth follow-up scans. Additionally, we employ a pre-trained age predictor to further validate the realism of the synthesized images with respect to the expected aging patterns.

Our results show that NeuroAR effectively learns aging-related morphological changes in subjects. By enhancing the accuracy of age-conditioned brain synthesis, our framework has significant implications for neurodegenerative disease modeling, longitudinal studies, and data augmentation in deep learning applications.

\section{Methods}
Our model, NeuroAR, is a novel 3D brain aging simulation model that employs a scale-wise autoregressive transformer paradigm for predicting the future brain scan of a subject based on his/her previous scan. It leverages the recent breakthrough of autoregressive transformers in the language domain \cite{gpt_series_3,ar_llms_1,ar_llms_9_team2023gemini}
by tailoring an autoregressive transformer structure for this task. NeuroAR enables conditional volumetric synthesis of a full brain MRI with fast inference speed and high fidelity.

\subsection{Problem Formulation}
Brain aging synthesis involves generating a structurally plausible future MRI scan of a subject’s brain given an earlier scan, the age at which it was acquired, and the target age. Formally, let $\mathbf{X}_t$ represent a brain MRI scan acquired at age t. The goal is to learn a function f such that 

  \begin{align}
      \hat{\mathbf{X}}_{t+\Delta t} = f(\mathbf{X}_t, t, t+\Delta t)
  \end{align}  

where $\hat{\mathbf{X}}_{t+\Delta t}$ is the predicted brain scan at the future age $t+\Delta t$. The function f should model the complex morphological transformations associated with aging while preserving subject-specific characteristics.

\subsection{Autoencoder}
To construct a convenient multi-scale latent space, we first train a VQVAE. Receiving a 3D MR image $x \in \mathbb{R}^{H\times W\times D}$, the encoder maps its input onto discrete token maps $\mathbf{f}_1, \mathbf{f}_2, ..., \mathbf{f}_S$ across S spatial scales. To do this, the encoder first extracts a continuous latent representation $z \in \mathbb{R}^{h_S x w_S x d_S x c}$ ($h_S$, $w_S$, $d_S$ denoting spatial dimensions at the Sth scale-the highest scale) via a set of residual convolutional blocks and a middle attention layer: $z=E(\mathbf{x})$. Afterwards, the discrete token maps at the sth scale are extracted by spatial downsampling to attain dimensions of $h_s x w_s x d_s$ followed by a quantization procedure based on a learnable codebook $B \in \mathbb{R}^{Vxc}$ with vocabulary size V. B characterizes a discrete latent space comprising V categories.

To minimize redundancy and information losses across scales, we adopt a hierarchical procedure for token map extraction that progresses from the lowest to the
highest spatial scale. For this purpose, a residual continuous representation $r_s \in \mathbb{R}^{h_S x w_S x d_S x c}$ is maintained, initialized as $\mathbf{r}_1=z$ at $s = 1$. At the sth scale, the residual continuous representation is used to derive $\mathbf{f}_s$ as follows:

\[
\mathbf{f}_s = \argmin_{v \in \{1, \dots, V\}} \left\| \mathbf{B}(v, :) - \text{D}_s(\mathbf{r}_s) \right\|_2^2,
\]
where $\mathbf{f}_s \in [V]^{h_s x w_s x d_s}$, D$_s$ denotes spatial downsampling to the sth scale via interpolation, and quantization is attained by identifying the closest vector in the codebook according to Euclidean distance.  Afterwards, codebook vectors corresponding to $\mathbf{f}_s$ are retrieved and upsampled via interpolation to the Sth scale, and used to update the residual representation:
\begin{equation}
    \mathbf{r}_{s+1} = \mathbf{r}_s - \text{Conv}(\text{Up}_S(\text{Lookup}(\mathbf{B},\mathbf{f}_s)))
    \label{dec_res}
\end{equation}
where Lookup is the retrieval function for codebook vectors given discrete token maps, Up$_S$ denotes spatial upsampling to the Sth scale via interpolation, and Conv denotes a convolutional layer.

On the other hand, the decoder maps the discrete multi-scale token maps ${\mathbf{f}_1, \mathbf{f}_2, ..., \mathbf{f}_S}$ back to the corresponding MR image $\mathbf{x}$ from which they were derived. The decoder starts with $\hat{\mathbf{r}}_0=0 \in \mathbf{R}^{h_S x w_S x d_S x c}$. At the sth scale, codebook vectors corresponding to the discrete token map $\mathbf{f}_s$ are retrieved and upsampled and used to update the predictions for the residual representation as follows:
\begin{equation}
    \hat{\mathbf{r}}_s = \hat{\mathbf{r}}_{s-1} + \text{Conv}(\text{Up}_S(\text{Lookup}(\mathbf{B}, \mathbf{f}_s))).
\end{equation}

Then, the predicted continuous representation at the highest scale \(  \hat{\mathbf{r}}_S  \) is used to recover the original image via projection through a residual convolutional network: $ \hat{\mathbf{x}} = D(\hat{\mathbf{r}}_S)$ where \( \hat{\mathbf{x}} \in \mathbb{R}^{H \times W \times D} \) denotes a prediction of the original MR image input to the VAE encoder. The autoencoder is trained via the following objective: 
\begin{align*}
 \mathcal{L}  =&  \lambda_{L_1} \lvert \mathbf{x} - \hat{\mathbf{x}} \rvert_1 
    + \lambda_p \sum_{p=1}^{m} \lvert V_p(\mathbf{x}) - V_p(\hat{\mathbf{x}}) \rvert_1 
      + \lambda_q \sum_{k=1}^{S}
    ( \left(z - \text{sg}[\hat{\mathbf{r}}_k] \right)^2 + \beta \left(\text{sg}[z]  - \hat{\mathbf{r}}_k \right)^2 ) \\ &+ \lambda_G \mathcal{L}_{adv} \left( \hat{\mathbf{x}} \right)
\end{align*}
where $\lambda_{L_1}$, $\lambda_p$, $\lambda_q$ and $\lambda_{adv}$ denotes the coefficients of the $L_1$, perceptual, quantization and adversarial losses, respectively. $V_p$ for $p=1,...,m$ denotes a set of layers of the pretrained VGG model \cite{simonyan2015vgg}. Following \cite{vqvae}, quantization loss was implemented as the sum of codebook loss and commitment loss using stopgradient operator denoted as 'sg'. A discriminator was trained together with autoencoder to ensure realistic reconstruction via adversarial learning.   

\subsection{Neural Autoregressive Model}

\begin{figure}
\includegraphics[width=\textwidth]{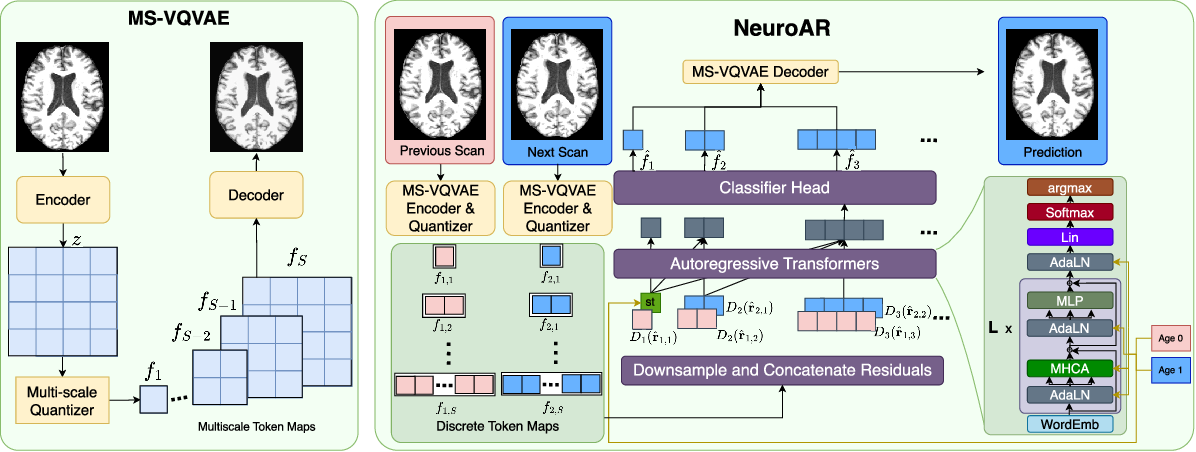}
\caption{Architecture of the NeuroAR model. \textbf{Left:} Core component of NeuroAR, which is a multi-scale vector-quantized variational autoencoder (MS-VQVAE). Its encoder module maps an MR scan to discrete token maps across $S$ spatial scales, and its decoder module recovers the input MR scan back from the derived token maps. \textbf{Right:} NeuroAR first encodes two MRI scans into tokens and concatenates them, doubling the word length. Then, a transformer module is employed to generate multi-scale tokens in an autoregressive fashion across multiple spatial scales.  To guide the generation via the previous and the next age, age embeddings are used to construct a start token and join each transformer block via cross-attention.} \label{fig1}
\end{figure}

The architecture of NeuroAR is depicted in Fig. \ref{fig1}. To start the generation, we first extract the discrete token maps of both the previously acquired MR scan and the MR scan to be predicted (denoted as the next scan in Fig. \ref{fig1}. We use $\mathbf{f}_{1,1}, \mathbf{f}_{1,2}, ..., \mathbf{f}_{1,S}$ to denote the discrete token maps of the previous scan and  $\mathbf{f}_{2,1}, \mathbf{f}_{2,2}, ..., \mathbf{f}_{2,S}$to denote the discrete token maps of the next scan. We then generate the residuals as in equation \ref{dec_res}. We denote the residuals of the previous scan as $\hat{\mathbf{r}}_{1,1}, \hat{\mathbf{r}}_{1,2}, ... , \hat{\mathbf{r}}_{1,S}$ and the residuals of the next scan as $\hat{\mathbf{r}}_{2,1}, \hat{\mathbf{r}}_{2,2}, ... , \hat{\mathbf{r}}_{2,S}$. We then downsample the nth residual of the previous scan to the nth scale for n=1,...,S, denoted as $D_n(\hat{\mathbf{r}}_{1,n})$. Also, we downsample the nth residual of the next scan to the (n+1)th scale for n=0,...,S-1, denoted as $D_{n+1}(\hat{\mathbf{r}}_{2,n})$. At the first scale, we concatenate $D_1(\hat{\mathbf{r}}_{1,1})$ with the start token calculated from the age embeddings. For $s \neq 1$, on the sth scale, we concatenate $D_n(\hat{\mathbf{r}}_{1,n})$ with $D_{n+1}(\hat{\mathbf{r}}_{2,n})$. During training, we employ teacher-forcing; whereas, during inference, the residuals of the next scan are autoregressively computed based on the output of the transformer.

Our transformer architecture starts with the word embedding layer, which uses a linear layer to extract embeddings from the given residuals. Then, we have a repeated set of L blocks as in Fig. \ref{fig1}. In our experiments, we use L=32 blocks. The Adaptive Normalization Layer (AdaLN) applies layer normalization and then induces age-guided processing of its input h as follows:
\begin{equation}
    \text{AdaLN}(h,\text{Age}_0,\text{Age}_1) = \text{LN}(h)\phi_{scale}(\text{Age}_0,\text{Age}_1) + \phi_{shift}(\text{Age}_0,\text{Age}_1)    
\end{equation}
where LN denotes Layer Normalization, and $\phi_{scale}$, $\phi_{shift}$ denote linear layers to compute scaling and shifting parameters based on the previous and target age. MHCA stands for multi-head cross-attention. To apply cross-attention here, we compute the query and value from the inputs to the MHCA layer and the key from the ages. MLP stands for multi-layer perceptron and is used at the end of each block. At the end of NeuroAR, we employ the Classifier Head to predict the indices of the tokens of the next scan. NeuroAR is trained with the Cross-Entropy loss.

\subsection{Competing Methods}

\textbf{LDM} (Latent Diffusion Model): As a competing method, we employ an LDM \cite{rombach2022latentdiffusion} to generate the future MR scan at a desired age. We utilize the MONAI framework for this method \cite{pinaya2023generativeaimedicalimaging}. We condition the latent diffusion model on the previous scan's latent representation, the age at which the previous scan was acquired, as well as the target age.

\noindent \textbf{Latent StarGAN}: As another competing method, we adapt StarGAN \cite{choi2018starganunifiedgenerativeadversarial} for our task. To ensure memory compliance in 3D volumetric training, we train StarGAN on the latent space of an autoencoder.  We use the implementation from \cite{Wongtrakool2025SyntheticCT} as a base. Instead of using the discriminator network for domain classification as in \cite{choi2018starganunifiedgenerativeadversarial}, we use a pretrained age regressor network from Peng et al. \cite{Peng2021AccurateBrainAge} as an age regressor in the objective of the StarGAN. Performance differences among competing methods were examined via nonparametric Wilcoxon signed-rank tests (p<0.05).

\section{Experiments}

\subsection{Dataset}
We evaluate NeuroAR using the ground-truth data on the ADNI (Alzheimer’s Disease Neuroimaging Initiative), PPMI (Parkinson’s Progression Markers Initiative) and ABCD (Adolescent Brain Cognitive Development Study) datasets. Since our objective in this work is to model the aging on control subjects, we only use the control subjects. Resultantly, on ADNI and PPMI, we employ a dataset of 1258 images from subjects aged 31 to 95. On ABCD, we use 21008 images from subjects aged from 107 months (approximately 8.92 years) to 189 months (15.75 years). As preprocessing, we have skull-stripped each image using FreeSurfer and then registered it to the MNI152 template by ANTs (Advanced Normalization Tools (ANTs)).

\subsection{Implementation details}
Throughout our experiments, we keep (H,W,D)=(160,192,176) as the image size, as it is our template size. We use S=5 spatial scales, which are $(h_1,w_1,d_1)=(2,2,2)$, $(h_2,w_2,d_2)=(4,4,4)$, $(h_3,w_3,d_3)=(6,8,7)$, $(h_4,w_4,d_4)=(8,10,9)$, $(h_5,w_5,d_5)=(10,12,11)$. We use V=2048 learnable codebook vectors throughout our experiments. To train the MS-VQVAE, we use a downsampling rate of 16 at each dimension due to memory constraints. The autoregressive transformer module employed L=32 sequential blocks with embedding dimensionality of 1024 and 32 cross-attention heads. It was trained with Adam \cite{kingma2015adam} optimizer. We also employ early stopping. Similarly to the VQ-VAE used in NeuroAR, we use a downsampling factor of 16 from each dimension to train the autoencoder for the LDM. We used 1000 denoising steps in our LDM. Also, on our latent StarGAN, we use the same autoencoder as the LDM's autoencoder.

\subsection{Evaluations}
We evaluated the fidelity of the proposed NeuroAR model's synthesis and compared it with the state-of-the-art methods, including the latent diffusion model (LDM) and latent StarGAN. As seen in Table \ref{tab:results}, NeuroAR consistently achieves the highest image fidelity, as manifested by the higher PSNR and SSIM scores on both aging adults and children. Nonparametric Wilcoxon signed-rank tests indicate that NeuroAR reached significantly (p<0.05) better generation than LDM and latent StarGAN on all the datasets with respect to both PSNR and SSIM. 

\begin{table}[h]
    \centering
    \caption{Accuracy of longitudinal image synthesis using known ground-truth data on different datasets using NeuroAR, LDM, and L-StarGAN}
    \scalebox{0.85}{\begin{tabular}{l c c c c c c}
        \toprule
         & \multicolumn{2}{c}{\textbf{ADNI}} & \multicolumn{2}{c}{\textbf{PPMI}} & \multicolumn{2}{c}{\textbf{ABCD}}\\
        \cmidrule(lr){2-3} \cmidrule(lr){4-5} \cmidrule(lr){6-7}
        & \textbf{PSNR} & \textbf{SSIM} & \textbf{PSNR} & \textbf{SSIM} & \textbf{PSNR} & \textbf{SSIM} \\
        \midrule
        \textbf{NeuroAR} & \textbf{21.08 (1.24)} & \textbf{0.837} \textbf{(0.025)} & \textbf{21.63} \textbf{(0.46)} & \textbf{0.827} \textbf{(0.008)} & \textbf{22.60 (0.83)} & \textbf{0.836 (0.016)} \\
        \textbf{LDM} &18.10 (0.89)& 0.714 (0.019)&  18.95(0.44)& 0.714 (0.006) & 19.03 (0.45) & 0.674 (0.013)  \\
        
        \textbf{L-StarGAN} & 16.35 (0.79) & 0.633 (0.018) & 17.14 (0.57) & 0.640 (0.015) & 18.87 (0.73) & 0.671 (0.016)\\
        \bottomrule
    \end{tabular}
    }
    \label{tab:results}
\end{table}

\begin{figure}
\includegraphics[width=\textwidth]{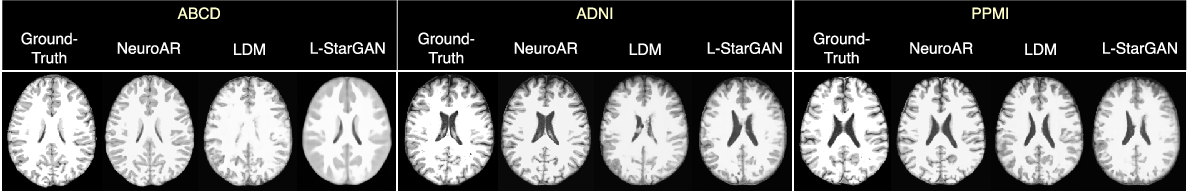}
\caption{Representative synthesized images by NeuroAR, LDM, and L-StarGAN on ABCD, ADNI, and PPMI datasets together with the ground-truth image.} \label{images}
\end{figure}

Sample synthetic MR images from each method on ABCD, ADNI and PPMI datasets are shown in Fig. \ref{images}. They showcase that images from NeuroAR preserve better anatomical structure and accurate details. 

To further validate the strength of the autoregressive longitudinal modeling approach, we generated synthetic MR images on the ABCD dataset. To generate the images, we determined the desired age of generation for each subject as the age from the last scan plus 2. We then merged this synthesized data with the real ABCD data. Then, following Peng et. al.\cite{Peng2021AccurateBrainAge}, we trained a CNN for age prediction on both the mixed data and the real data. We used the same architecture as in \cite{Peng2021AccurateBrainAge} for both models and the same output age bins. To not change the maximum age in the training set, we did not include synthetic images exceeding the ABCD dataset's maximum age, and we did not include synthetic images for the fourth session. Resultantly, the real train set contained 18,879 images while the mixed train set contained 26212 images, meaning a 38.8\% increase by augmentation. Table \ref{tab:age_regression_abcd} shows the mean absolute error and R\textsubscript{2} scores on real data.

\begin{table}[t]
    \centering
    \caption{Performances of two age regression models on real data on the ABCD Dataset. MAE denotes mean absolute error in age prediction. The top model was trained with synthesized scans and outperforms the bottom model on MAE and R\textsubscript{2} scores.}
    \renewcommand{\arraystretch}{1.3} 
    \begin{tabular}{l c c}
        \toprule
        \textbf{Training Strategy~~~} & \textbf{~MAE~} & \textbf{~~R\textsuperscript{2}~~} \\
        \midrule
        {Real Data only} & 0.685& 0.807  \\
        {Real \& Synthetic} & \textbf{0.582} & \textbf{0.836} \\
        \bottomrule
    \end{tabular}
    \label{tab:age_regression_abcd}
\end{table}

\section{Conclusion}
In this study, we introduced NeuroAR, a novel autoregressive transformer-based model for brain aging synthesis. By leveraging a concatenated longitudinal token strategy and age guidance via cross attention and adaptive layer normalization, NeuroAR successfully models subject-specific brain aging trajectories while preserving anatomical consistency. Our evaluation across ADNI, PPMI, and ABCD datasets demonstrates that NeuroAR significantly outperforms latent diffusion models (LDM) and Latent StarGAN in terms of image fidelity, as reflected in higher PSNR and SSIM scores. Furthermore, we showed that incorporating NeuroAR-generated data in the age regression task improves model performance, reducing MAE and increasing R\textsuperscript{2} scores.

\section*{Acknowledgment}
This work was supported in part by National Institutes of Health under grants AG089169 and 4R00LM01430903 and the Stanford Institute for Human-Centered AI (HAI) Hoffman-Yee Award. 

\newpage
\bibliographystyle{splncs04}
\bibliography{references}

\begin{thebibliography}{10}
\providecommand{\url}[1]{\texttt{#1}}
\providecommand{\urlprefix}{URL }
\providecommand{\doi}[1]{https://doi.org/#1}

\bibitem{gpt_series_3}
Brown, T., Mann, B., Ryder, N., Subbiah, M., Kaplan, J., Dhariwal, P., Neelakantan, A., Shyam, P., Sastry, G., Askell, A., et~al.: Language models are few-shot learners. Advances in Neural Information Processing Systems  \textbf{33},  1877--1901 (2020)

\bibitem{choi2018starganunifiedgenerativeadversarial}
Choi, Y., Choi, M., Kim, M., Ha, J.W., Kim, S., Choo, J.: Stargan: Unified generative adversarial networks for multi-domain image-to-image translation. In: 2018 IEEE/CVF Conference on Computer Vision and Pattern Recognition. pp. 8789--8797 (2018). \doi{10.1109/CVPR.2018.00916}

\bibitem{ar_llms_1}
Chowdhery, A., Narang, S., Devlin, J., Bosma, M., Mishra, G., Roberts, A., Barham, P., Chung, H., Sutton, C., Gehrmann, S., et~al.: Palm: Scaling language modeling with pathways. Journal of Machine Learning Research  \textbf{24}(240),  1--113 (2023)

\bibitem{huizinga2018kernel}
Huizinga, W., Poot, D., Vernooij, M., Roshchupkin, G., Bron, E., Ikram, M., Rueckert, D., Niessen, W., Klein, S.: A spatio-temporal reference model of the aging brain. Neuroimage  \textbf{169},  11--22 (2018)

\bibitem{kingma2015adam}
Kingma, D.P., Ba, J.: Adam: A method for stochastic optimization. In: International Conference on Learning Representations (ICLR) (2015), \url{https://arxiv.org/abs/1412.6980}

\bibitem{lorenzi2015linear}
Lorenzi, M., Pennec, X., Frisoni, G., Ayache, N., Initiative, A.: Disentangling normal aging from alzheimer's disease in structural magnetic resonance images. Neurobiol. Aging  \textbf{36},  S42--S52 (2015)

\bibitem{modat2014nonrigid}
Modat, M., Simpson, I., Cardoso, M., Cash, D., Toussaint, N., Fox, N., Ourselin, S.: Simulating neurodegeneration through longitudinal population analysis of structural and diffusion weighted mri data. In: MICCAI. pp. 57--64. Springer (2014)

\bibitem{vqvae}
van~den Oord, A., Vinyals, O., Kavukcuoglu, K.: Neural discrete representation learning. In: Proceedings of the 31st International Conference on Neural Information Processing Systems. p. 6309–6318. NIPS'17, Curran Associates Inc., Red Hook, NY, USA (2017)

\bibitem{Peng2021AccurateBrainAge}
Peng, H., Gong, W., Beckmann, C.F., Vedaldi, A., Smith, S.M.: Accurate brain age prediction with lightweight deep neural networks. Medical Image Analysis  \textbf{68},  101871 (February 2021). \doi{10.1016/j.media.2020.101871}, \url{https://www.sciencedirect.com/science/article/pii/S1361841520302358}

\bibitem{pinaya2023generativeaimedicalimaging}
Pinaya, W.H.L., Graham, M.S., Kerfoot, E., Tudosiu, P.D., Dafflon, J., Fernandez, V., Sanchez, P., Wolleb, J., da~Costa, P.F., Patel, A., Chung, H., Zhao, C., Peng, W., Liu, Z., Mei, X., Lucena, O., Ye, J.C., Tsaftaris, S.A., Dogra, P., Feng, A., Modat, M., Nachev, P., Ourselin, S., Cardoso, M.J.: Generative ai for medical imaging: extending the monai framework (2023), \url{https://arxiv.org/abs/2307.15208}

\bibitem{pombo2023equitable}
Pombo, G., Gray, R., Cardoso, M.J., Ourselin, S., Rees, G., Ashburner, J., Nachev, P.: Equitable modelling of brain imaging by counterfactual augmentation with morphologically constrained 3d deep generative models. Medical Image Analysis  \textbf{84},  102723 (2023)

\bibitem{puglisi2025brainlatent}
Puglisi, L., Alexander, D.C., Ravì, D.: Brain latent progression: Individual-based spatiotemporal disease progression on 3d brain mris via latent diffusion. arXiv preprint  (2025), \url{https://arxiv.org/abs/2502.08560}

\bibitem{ravi2019daninetmiccai}
Ravi, D., Alexander, D.C., Oxtoby, N.P.: Degenerative adversarial neuroimage nets: Generating images that mimic disease progression. In: Shen, D., Liu, T., Peters, T.M., Staib, L.H., Essert, C., Zhou, S., Yap, P.T., Khan, A. (eds.) Medical Image Computing and Computer Assisted Intervention -- MICCAI 2019. pp. 164--172. Springer International Publishing, Cham (2019). \doi{10.1007/978-3-030-32248-9_19}

\bibitem{ravi2022degenerative}
Ravi, D., Blumberg, S.B., Ingala, S., Barkhof, F., Alexander, D.C., Oxtoby, N.P.: Degenerative adversarial neuroimage nets for brain scan simulations: Application in ageing and dementia. Medical Image Analysis  \textbf{75},  102257 (2022). \doi{10.1016/j.media.2021.102257}, \url{https://www.sciencedirect.com/science/article/pii/S1361841521003029}

\bibitem{rombach2022latentdiffusion}
Rombach, R., Blattmann, A., Lorenz, D., Esser, P., Ommer, B.: High-resolution image synthesis with latent diffusion models. arXiv preprint  (2022), \url{https://arxiv.org/abs/2112.10752}

\bibitem{serag2012kernel}
Serag, A., Aljabar, P., Ball, G., Counsell, S., Boardman, J., Rutherford, M., Edwards, A., Hajnal, J., Rueckert, D.: Construction of a consistent high-definition spatio-temporal atlas of the developing brain using adaptive kernel regression. Neuroimage  \textbf{59}(3),  2255--2265 (2012)

\bibitem{sharma2010nonrigid}
Sharma, S., Noblet, V., Rousseau, F., Heitz, F., Rumbach, L., Armspach, J.P.: Evaluation of brain atrophy estimation algorithms using simulated ground-truth data. Med Image Anal  \textbf{14}(3),  373--389 (2010)

\bibitem{simonyan2015vgg}
Simonyan, K., Zisserman, A.: Very deep convolutional networks for large-scale image recognition. arXiv preprint  (2015), \url{https://arxiv.org/abs/1409.1556}

\bibitem{sivera2019linear}
Sivera, R., Delingette, H., Lorenzi, M., Pennec, X., Ayache, N.: A model of brain morphological changes related to aging and alzheimer’s disease from cross-sectional assessments. Neuroimage  \textbf{198},  255--270 (2019)

\bibitem{ar_llms_9_team2023gemini}
Team, G., Anil, R., Borgeaud, S., Wu, Y., Alayrac, J.B., Yu, J., Soricut, R., Schalkwyk, J., Dai, A.M., Hauth, A., et~al.: Gemini: A family of highly capable multimodal models. arXiv preprint  (2023)

\bibitem{VAR}
Tian, K., Jiang, Y., Yuan, Z., PENG, B., Wang, L.: Visual autoregressive modeling: Scalable image generation via next-scale prediction. In: The Thirty-eighth Annual Conference on Neural Information Processing Systems (2024), \url{https://openreview.net/forum?id=gojL67CfS8}

\bibitem{Wongtrakool2025SyntheticCT}
Wongtrakool, P., Puttanawarut, C., Changkaew, P., Piasanthia, S., Earwong, P., Stansook, N., Khachonkham, S.: Synthetic ct generation from cbct and mri using stargan in the pelvic region. Radiation Oncology  \textbf{20}(1), ~18 (February 2025). \doi{10.1186/s13014-025-02590-2}, \url{https://ro-journal.biomedcentral.com/articles/10.1186/s13014-025-02590-2}

\bibitem{xia2019brainage}
Xia, T., Chartsias, A., Tsaftaris, S.A.: Consistent brain ageing synthesis. In: Shen, D., Liu, T., Peters, T.M., Staib, L.H., Essert, C., Zhou, S., Yap, P.T., Khan, A. (eds.) Medical Image Computing and Computer Assisted Intervention -- MICCAI 2019. pp. 750--758. Springer International Publishing, Cham (2019). \doi{10.1007/978-3-030-32251-9_82}

\bibitem{yoon2023sadm}
Yoon, J.S., Zhang, C., Suk, H.I., Guo, J., Li, X.: Sadm: Sequence-aware diffusion model for longitudinal medical image generation. In: Information Processing in Medical Imaging (2023)

\bibitem{zhang2016patch}
Zhang, Y., Shi, F., Wu, G., Wang, L., Yap, P.T., Shen, D.: Consistent spatial-temporal longitudinal atlas construction for developing infant brains. TMI  \textbf{35}(12),  2568--2577 (2016)

\bibitem{Ziegler_et_al}
Ziegler, G., Dahnke, R., Gaser, C., Initiative, A.D.N.: Models of the aging brain structure and individual decline. Frontiers in Neuroinformatics  \textbf{6}, ~3 (2012). \doi{10.3389/fninf.2012}

\end{thebibliography}

\end{document}